\documentclass{article}

\usepackage{PRIMEarxiv}

\usepackage[utf8]{inputenc} 
\usepackage[T1]{fontenc}    
\usepackage{hyperref}       
\usepackage{url}            
\usepackage{booktabs}       
\usepackage{amsfonts}       
\usepackage{nicefrac}       
\usepackage{microtype}      
\usepackage{lipsum}
\usepackage{fancyhdr}       
\usepackage{graphicx}       
\graphicspath{{media/}}     
\usepackage[table,xcdraw]{xcolor}

\usepackage{tabularx}      
\usepackage{colortbl}      
\usepackage[font=small]{caption}  
\usepackage{amsmath}

\newtheorem{theorem}{Theorem}[section]
\newtheorem{proposition}[theorem]{Proposition}
\newtheorem{lemma}[theorem]{Lemma}
\newtheorem{corollary}[theorem]{Corollary}

\newtheorem{assumption}[theorem]{Assumption}

\newtheorem{proof}[theorem]{Proof}

\usepackage{enumerate}
\usepackage{algorithm}
\usepackage{algorithmic}

\title{Efficient Generalization via Multimodal Co-Training \\ under Data Scarcity and Distribution Shift
}

\author{
  Tianyu Bell Pan, Damon L. Woodard\\
  Department of Electrical and Computer Engineering \\
  Florida Institute for National Security (FINS) \\
  Applied Artificial Intelligence Group \\
  University of Florida \\
  Gainesville, FL, 32611\\
  \texttt{\{tpan1, dwoodard\}@ufl.edu} \\
}

\begin{document}
\maketitle

\begin{abstract}
This paper explores a multimodal co-training framework designed to enhance model generalization in situations where labeled data is limited and distribution shifts occur. We thoroughly examine the theoretical foundations of this framework, deriving conditions under which the use of unlabeled data and the promotion of agreement between classifiers for different modalities lead to significant improvements in generalization. We also present a convergence analysis that confirms the effectiveness of iterative co-training in reducing classification errors. In addition, we establish a novel generalization bound that, for the first time in a multimodal co-training context, decomposes and quantifies the distinct advantages gained from leveraging unlabeled multimodal data, promoting inter-view agreement, and maintaining conditional view independence. Our findings highlight the practical benefits of multimodal co-training as a structured approach to developing data-efficient and robust AI systems that can effectively generalize in dynamic, real-world environments. The theoretical foundations are examined in dialogue with, and in advance of, established co-training principles.
\end{abstract}


\section{Introduction}
The ability to generalize flexibly and efficiently from limited resources is a key feature of human intelligence~\cite{griffiths2020understanding}. In contrast, artificial intelligence (AI) models struggle to expand their capabilities when training data and computational resources are scarce, hindering their effective deployment in real-world scenarios that demand learning from sparse samples in unpredictable environments~\cite{gao2024multi,wang2024multi}. A primary challenge is the scarcity of labeled data, which is often difficult and costly to obtain, especially in specialized or rapidly evolving fields~\cite{li2024learning,wang2022open}. This reliance on supervised learning reduces the robustness and generalizability of models. Additionally, most of the real-world data is multimodal, originating from diverse sources. While integrating these modalities can enhance understanding and robustness, many existing methods favor single-modality approaches or necessitate substantial labeled multimodal data for effective fusion~\cite{li2024multimodal}. 

Furthermore, AI models face non-stationary inputs and distribution shifts, which can degrade performance as learned patterns from historical data become outdated~\cite{liu2017regional,lu2018learning}. Such shifts are not mere academic concerns but practical hurdles, as seen in the evolving nature of misinformation campaigns~\cite{bayram2022concept} or the performance degradation of even large vision-language models when confronted with discrepancies between training and testing domains~\cite{wu2025multi}. Adversarial threats also evolve, introducing further complexity~\cite{shyaa2024evolving}.

This paper explores a semi-supervised multimodal co-training framework to tackle the challenge of limited labeled data. By leveraging unlabeled data and ensuring consistency across different modalities, AI systems can enhance their generalization capabilities, even in the face of label scarcity or distribution shifts. We analyze the mathematical foundations underlying this framework, investigating the convergence properties of the multimodal co-training algorithm and deriving a tighter generalization bound that highlights the advantages of using unlabeled data. These theoretical results highlight the potential of unlabeled multimodal data to address the scarcity of labeled data, thereby paving the way for the development of more data-efficient and adaptive AI systems.

\subsection{Contributions}
The key contributions of this research are:

\textbf{Multimodal co-training:} We introduce a theoretically grounded multimodal co-training framework that integrates dual-threshold pseudo-labeling, a customizable agreement loss, and a controlled label-expansion budget. These components are not ad-hoc but are designed to directly optimize factors identified in our theoretical analysis, such as improving pseudo-label reliability to aid convergence and shaping the agreement term $\Gamma$ in the generalization bound. The framework aims to optimize the analytical mixing coefficient $\alpha$ and reduce the contraction factor $\lambda$. 

\textbf{Convergence guarantee:} We provide a rigorous proof of geometric convergence for the iterative co-training process, demonstrating that classifier error is reduced in expectation under standard co-training assumptions. This analysis clarifies the conditions for stable learning and the role of inter-view information exchange in error reduction, moving beyond heuristic justifications for the efficacy of co-training. Each round of pseudo-label exchange and agreement loss significantly reduces classifier error in expectation, resulting in geometric convergence up to a minimum.

\textbf{Novel generalization bound:} We derive a novel PAC-style generalization risk bound that isolates and quantifies a benefit term ($\Gamma$). This term captures how the fraction of unlabeled data, the degree of inter-view agreement, and the strength of view on conditional independence collectively tighten the bound. This provides a more interpretable understanding of the value of unlabeled multimodal data in co-training than prior guarantees, surpassing prior co-training and semi-supervised guarantees by offering this decomposition. 

\textbf{Data-efficient learning:} We show that increasing the amount of unlabeled data or enhancing view independence and agreement can significantly improve both convergence and generalization. These findings offer practical strategies for leveraging large corpora of unlabeled data and promoting consistency to address label scarcity and distribution shifts.

\subsection{Related Work}
Research on learning with limited annotated data encompasses several areas of study. Few-shot and zero-shot learning (FSL/ZSL) methods aim to generalize from very few or no labeled examples by leveraging auxiliary information, such as semantic attributes or meta-learning~\cite{chamieh2024llms}. However, these methods often face challenges in high-dimensional or multimodal settings and can be biased when attribute representations are too general and non-discriminative~\cite{wanyan2024comprehensive,liu2025boosting,tang2024semantic}. Simultaneously, multimodal fusion approaches integrate various types of inputs (e.g., text, images, audio) through early, intermediate, or late fusion strategies~\cite{li2024multimodal,xu2013survey}. However, these approaches typically require large labeled datasets to effectively align and combine different modalities, making them fragile when labeled data is scarce~\cite{yu2025review}.

Furthermore, semi-supervised learning (SSL) methods, particularly co-training, utilize unlabeled data by training separate classifiers on conditionally independent "views" and exchanging high-confidence pseudo-labels~\cite{blum1998combining}. However, existing analyses of SSL mainly focus on single-view scenarios or provide only loose additive guarantees without clearly quantifying how agreement between views reduces risk. Additionally, open-world deployment requires models to effectively generalize under non-stationary data distributions, where distribution shifts can significantly degrade performance~\cite{lu2018learning}. While some approaches utilize drift detection mechanisms to initiate model updates, it is important to understand the robustness of different learning paradigms in response to such shifts~\cite{bayram2022concept}. 

To address these challenges, we propose a unified framework that integrates multiple modalities and unlabeled data while ensuring rigorous convergence and risk bounds. Our proposed multimodal co-training framework introduces a novel approach that utilizes at least two different modalities and leverages unlabeled data. Instead of relying solely on the direct integration of representations derived from limited labeled data, the multimodal co-training framework utilizes unlabeled data along with the principle of cross-view consistency~\cite{li2024multimodal}. This method enhances and regularizes the specific views of individual learners before any final integration, resulting in more refined features.

\section{Preliminary and Problem Formulation}
\subsection{Notation and Definitions}
Let $\mathcal{X}$ denote the input space and $\mathcal{Y}$ be the label space. For binary classification, we consider $\mathcal{Y}=\{0,1\}$ or $\mathcal{Y}=\{-1,+1\}$. Data instances $(x, y)$ are assumed to be drawn from a true underlying data distribution $P$. This distribution may be non-stationary in open-world scenarios, denoted as $P_t(x,y)$ at a given time $t$. We are provided with a set of $N_L$ labeled examples, $\mathcal{L}_{\text{labeled}}=\left\{\left(x_i, y_i\right)\right\}_{i=1}^{N_L}$, drawn from $P$. Additionally, we have access to a typically much larger set of $N_U$ unlabeled examples, $\mathcal{U}=\left\{u_j\right\}_{j=1}^{N_U}$, drawn from the marginal distribution $P_X$. A hypothesis (or classifier) is a function $h: \mathcal{X} \rightarrow \mathcal{Y}$. The quality of a hypothesis $h$ is measured by its true risk (generalization error) concerning the distribution $P$:
\begin{align}
    R(h)=\mathbb{E}_{(x,y) \sim P}[\ell(h(x),y)],
\end{align}
where $\ell(\cdot, \cdot)$ is a chosen loss function. The empirical risk of $h$ on the labeled data $\mathcal{L}_{labeled}$ is:
\begin{align}
    \hat{R}_{N_L}(h)=\frac{1}{N_L} \sum_{i=1}^{N_L} \ell\left(h\left(x_i\right), y_i\right).
\end{align}

Let $\mathcal{H}$ be the hypothesis class from which $h$ is chosen. The complexity of $\mathcal{H}$ can be characterized by measures such as the Vapnik-Chervonenkis (VC) dimension ($d_{VC}(\mathcal{H})$)~\cite{vapnik2015uniform}, or the Rademacher complexity ($R_N(\mathcal{H})$)~\cite{bartlett2002rademacher}.

\subsection{The Multimodal Learning Setting}
We consider a multimodal learning scenario where each input instance $x \in \mathcal{X}$ comprises representations from at least two distinct modalities. We focus on two views for clarity, but the framework can be extended. For example, $x=\left(x^{(1)}, x^{(2)}\right)$, where $x^{(1)} \in \mathcal{X}^{(1)}$ are features from the first modality (e.g., text features from a document) and $x^{(2)} \in \mathcal{X}^{(2)}$ are features from the second modality (e.g., image features from an associated image). We denote the view-specific feature extractors as $h^{(1)}: \mathcal{X}^{(1)} \rightarrow \mathcal{Y}$ and $h^{(2)}: \mathcal{X}^{(2)} \rightarrow \mathcal{Y}$.

In this setting, each modality $x^{(v)}$ can predict the label $y$ with reasonable accuracy. This means that there exist good classifiers $h^{*(1)}\left(x^{(1)}\right)$ and $h^{*(2)}\left(x^{(2)}\right)$. However, each view may also be noisy or incomplete; for example, text might be ambiguous, or an image might lack context. The premise is that the complementary information across views can be harnessed for improved learning. The core premise is that while each view is stochastically sufficient, their error patterns are diverse, allowing for mutual correction and enrichment through collaborative learning. This premise is that the complementary information across views can be utilized for improved learning. 

\subsection{Problem Formulation}
Given the following two sets of data: (1) a small set of labeled multimodal examples $\mathcal{L}_{\text{labeled}}=\left\{\left(\left(x_i^{(1)}, x_i^{(2)}\right), y_i\right)\right\}_{i=1}^{N_L}$; and (2) a large set of unlabeled multimodal examples $\mathcal{U}=\left\{\left(u_j^{(1)}, u_j^{(2)}\right)\right\}_{j=1}^{N_U}$. The objective is to learn a combined classifier, denoted $h_\theta\left(\left(x^{(1)}, x^{(2)}\right)\right)$, parameterized by $\theta$, that minimizes the true risk $R\left(h_\theta\right)$ on unseen future data drawn from $P$. This classifier $h_\theta$ typically involves view-specific components $h_{\theta_1}^{(1)}$ and $h_{\theta_2}^{(2)}$, and a mechanism for combining their outputs or enforcing agreement. This paper focuses on the theoretical properties (convergence and generalization) of achieving this objective via a multimodal co-training strategy.

\subsubsection{Co-Training Strategy in the Multimodal Context}
The co-training strategy involves two view-specific classifiers, $h^{(1)}$ (operating on $\mathcal{X}^{(1)}$) and $h^{(2)}$ (operating on $\mathcal{X}^{(2)}$). The process begins with initialization, where classifiers $h^{(1)}$ and $h^{(2)}$ are trained on the labeled set $\mathcal{L}_{\text{labeled}}$. This initial training sets the foundation for the subsequent iterative refinement steps.

In the next phase, labeled examples are further enhanced through iterative pseudo-labeling and retraining. Classifier $h^{(1)}$ predicts labels for examples in the unlabeled set $\mathcal{U}$. From these predictions, a confident subset $\mathcal{U}_{\text{pseudo}}^{(1)} \subset \mathcal{U}$ is selected, and these examples are assigned pseudo-labels $\hat{y}^{(1)}$. The pseudo-labeled instances $\left\{\left(u_j, \hat{y}_j^{(1)}\right)\right\}$ are then utilized to augment the training set for $h^{(2)}$. Conversely, classifier $h^{(2)}$ performs the same for $h^{(1)}$ by pseudo-labeling a confident subset $\mathcal{U}_{\text{pseudo}}^{(2)}$. Both $h^{(1)}$ and $h^{(2)}$ are subsequently retrained using the labeled set $\mathcal{L}_{\text{labeled}}$ along with their respective augmented pseudo-labeled datasets. This process is repeated for several rounds until a convergence criterion is satisfied.

To further leverage the unlabeled data and enforce consistency between the two classifiers, an agreement loss term ($\mathcal{L}_{\text{agree}}$) can be incorporated. This term penalizes discrepancies between the predictions of $h^{(1)}$ and $h^{(2)}$ on the same examples $u \in \mathcal{U}$. For a given unlabeled instance $u=\left(u^{(1)}, u^{(2)}\right)$, if $p^{(1)}(u)$ and $p^{(2)}(u)$ represent the probabilistic outputs from $h^{(1)}$ and $h^{(2)}$, then the agreement loss can be formulated as 
\begin{align}
    \mathcal{L}_{\text{agree}}=\mathbb{E}_{u \sim \mathcal{U}} D,
\end{align}
where $D$ is a divergence or distance measure, such as the squared difference $\left(p^{(1)}-p^{(2)}\right)^2$ or cross-entropy if one model's output serves as a target for the other. This loss term is typically integrated with the supervised loss during training, expressed as $\mathcal{L}_{\text{total}}=\mathcal{L}_{\text{sup}}+\lambda_{\text{agree}} \mathcal{L}_{\text{agree}}$, where $\mathcal{L}_{\text{sup}}$ indicates the empirical risk on the labeled set $\mathcal{L}_{\text{labeled}}$ and $\lambda_{\text{agree}}$ is a weighting factor that balances the two loss components. Additionally, mechanisms are implemented to regulate the rate of label expansion, which helps minimize the risk of error propagation throughout the training process.

\subsection{Proposed Algorithm}
Algorithm~\ref{alg:multimodal_cotraining_concise} is hence proposed. This algorithm trains separate classifiers on each view, encouraging mutual supervision through pseudo-label exchange and cross-view consistency. The key insight is that each classifier can help label examples for the other when confident, and both classifiers are encouraged to agree on predictions for unlabeled data. 

\begin{algorithm}[!ht]
  \caption{Multimodal Co‑Training}
  \label{alg:multimodal_cotraining_concise}
  \begin{algorithmic}[1]
    \REQUIRE Labeled set $\mathcal{L}=\{((x_i^{(1)},x_i^{(2)}),y_i)\}_{i=1}^{N_L}$; Unlabeled set $\mathcal{U}=\{(u_j^{(1)},u_j^{(2)})\}_{j=1}^{N_U}$; Iterations $K$; thresholds $\tau_{\mathrm{pseudo}},\tau_{\mathrm{agree}}$; pseudo‑label count $k_{\mathrm{pseudo}}$; agreement weight $\lambda_{\mathrm{agree}}$
    \ENSURE Trained models $h^{(1)}_{\theta_1},\,h^{(2)}_{\theta_2}$
    \STATE Initialize $h^{(1)}_{\theta_1},h^{(2)}_{\theta_2}$; set $\mathcal{U}_{\text{pool}}\gets\mathcal{U}$
    \FOR{$t = 1$ to $K$}
      \STATE $\mathcal{P}_1,\mathcal{P}_2\gets\emptyset$
      \FORALL{$(u^{(1)},u^{(2)})\in\mathcal{U}_{\mathrm{pool}}$}
        \FOR{$v = 1$ to $2$}
          \STATE $(p_u^{(v)},c_u^{(v)})\gets\mathrm{Predict}(h^{(v)}_{\theta_v},u^{(v)})$
          \IF{$c_u^{(v)}>\tau_{\mathrm{pseudo}}$}
            \STATE $\hat y_u^{(v)}\gets\mathrm{PseudoLabel}(p_u^{(v)})$; add $((u^{(1)},u^{(2)}),\hat y_u^{(v)},c_u^{(v)})$ to $\mathcal{P}_{3-v}$
          \ENDIF
        \ENDFOR
      \ENDFOR
      \STATE For each $v\in\{1,2\}$, select top‑$k_{\mathrm{pseudo}}$ by confidence in $\mathcal{P}_v$ as $\mathcal{L}_{\mathrm{pseudo}}^{(v)}$
      \STATE $\mathcal{D}^{(v)}\gets\mathcal{L}\cup\{(x,y)\mid (x,y,\cdot)\in\mathcal{L}_{\mathrm{pseudo}}^{(v)}\}$
      \FOR{$v = 1$ to $2$}
        \STATE Freeze $h^{(3-v)}_{\theta_{3-v}}$
        \STATE Update $\theta_v\gets\arg\min_{\theta}\Bigl[\sum\mathcal{L}_{\sup}+\lambda_{\mathrm{agree}}\sum\mathcal{L}_{\mathrm{agree}}\Bigr]$
      \ENDFOR
      \STATE Remove all $\mathcal{L}_{\mathrm{pseudo}}^{(1)}\cup\mathcal{L}_{\mathrm{pseudo}}^{(2)}$ from $\mathcal{U}_{\mathrm{pool}}$
      \IF{$\mathcal{L}_{\mathrm{pseudo}}^{(1)}\cup\mathcal{L}_{\mathrm{pseudo}}^{(2)}=\emptyset$}
        \STATE \textbf{break}
      \ENDIF
    \ENDFOR
    \STATE \textbf{return} $h^{(1)}_{\theta_1},\,h^{(2)}_{\theta_2}$
  \end{algorithmic}
\end{algorithm}

\section{Main Results}
This section presents the main results concerning the convergence and generalization of this proposed framework. 

\subsection{Assumptions}
To analyze the main framework, three assumptions are defined:
\begin{assumption}[Multi-View Structure and Sufficiency]\label{ass:1}
    The learning problem involves at least two views of the data, denoted $x^{(1)}$ and $x^{(2)}$, corresponding to different modalities. We assume that:
\begin{enumerate}[(i)]
    \item \textbf{View sufficiency:} Each view $x^{(v)}$, for $v \in\{1,2\}$, is individually sufficient for classification. Formally, for each view $v$, there exists a hypothesis $h^{*(v)}$ within the hypothesis space $\mathcal{H}^{(v)}$ such that its true risk $R\left(h^{*(v)}\right) \leq \epsilon_{\text{suff}}$, where $\epsilon_{\text{suff}}$ is a small positive constant.
    \item \textbf{Conditional independence of views:} Given the true label $y$, the features $x^{(1)}$ and $x^{(2)}$ are conditionally independent. That is, $P\left(x^{(1)}, x^{(2)}| y\right)=P\left(x^{(1)}|y\right) P\left(x^{(2)}|y\right)$.
\end{enumerate}
\end{assumption}

This assumption is foundational to the co-training paradigm. View sufficiency ensures that each modality-specific classifier has a learnable target and can, in principle, make accurate predictions. Conditional independence is critical for the efficacy of the co-training process. It implies that the errors made by one view are not systematically correlated with the errors of the other, given the true class. 

\begin{assumption}[Classifier Quality and Pseudo-Label Reliability]\label{ass:2}
    The co-training process effectively bootstraps performance based on initial classifier quality and the reliability of the information exchanged between views. We assume that:
    \begin{enumerate}[(i)]
    \item \textbf{Initial classifier quality:} The initial classifiers $h^{(v)}$, trained on the available labeled data $\mathcal{L}$, have an error rate $\epsilon_v$ on the unlabeled data distribution that is better than random change (i.e., $\epsilon_v < \frac{1}{2} - \delta_0$ for some small $\delta_0 > 0$).
    \item \textbf{Pseudo-label reliability:} During the co-training process, pseudo-labels $\tilde{y}^{(v)}$ generated by a classifier $h^{(v)}$ for confidently predicted unlabeled instances $u$ are correct with high probability.
\end{enumerate}
\end{assumption}

This assumption is that initial classifiers possess a baseline level of competence (i.e., better-than-random performance) and that the subsequent pseudo-labeling process maintains high reliability. The initial classifier quality is important because co-training is a bootstrapping mechanism. 

\begin{assumption}[Model Learnability and Loss Properties]\label{ass:3}
    To ensure that the learning process is well-posed and that generalization can be theoretically analyzed, we make standard assumptions regarding the model and loss function:
    \begin{enumerate}[(i)]
    \item \textbf{Bounded hypothesis complexity:} The hypothesis class $\mathcal{H}$ has bounded complexity. This can be characterized by a finite VC-dimension, $d_{VC}(\mathcal{H}) \leq d_{\text{eff}}$, or by Rademacher complexity $R_N(\mathcal{H})$ that grows sub-linearly with $N$.
    \item \textbf{Loss function properties:} The loss function $\ell(\cdot,\cdot)$ used for training is bounded.
\end{enumerate}
\end{assumption}

This assumption is a standard in statistical learning theory, ensuring the learning problem is well-posed and analyzable. Bounded hypothesis complexity prevents the models from becoming overly complex and overfitting the training data, including the initial labeled set and the incrementally added pseudo-labeled examples. This control over complexity is fundamental for deriving generalization bounds that guarantee performance on unseen data. Properties of the loss function ensure stability in risk calculations. 

\subsection{Convergence of Multimodal Co-Training}
Before presenting the convergence theorem, we introduce Lemma~\ref{lem1}, which describes the expected behavior of a classifier trained on pseudo-labels provided by a slightly more accurate view. 

\begin{lemma}[Expected Improvement via Pseudo‐Labeling]\label{lem1}
Let $h^{(1)}$ and $h^{(2)}$ be classifiers on two conditionally independent views with true error rates $\epsilon^{(1)}$ and $\epsilon^{(2)}$, respectively, where 
\begin{align}
    \epsilon^{(2)}<\min\bigl\{\epsilon^{(1)},\tfrac12\bigr\}.
\end{align}
Suppose $h^{(1)}$ is retrained on a sufficiently large set of pseudo‐labels produced by $h^{(2)}$, and that these pseudo‐labels are reliable under the conditional independence assumption. Then, there exists $\alpha\in(0,1]$ such that
\begin{align}
    \mathbb{E}\bigl[\epsilon'^{(1)}\bigr]\;\le\;(1-\alpha)\,\epsilon^{(1)}\;+\;\alpha\,\epsilon^{(2)}\;<\;\epsilon^{(1)}.
\end{align}
\end{lemma}

This lemma shows that a less accurate classifier can improve by learning from a more accurate one, given that the views are conditionally independent. Conditional independence prevents the pseudo-labels from $h^{(2)}$ from being systematically biased like $h^{(1)}$'s errors. If $h^{(2)}$ is more accurate and its pseudo-labels are mostly correct, $h^{(1)}$ can use these examples to correct its mistakes. The view sufficiency condition is important. A small number of pseudo-labels may not be sufficient to overcome existing biases, but a large set can drive effective learning. This single-step improvement is key to the iterative convergence of the co-training algorithm. Lastly, the expectation $\mathbb{E}[\cdot]$ captures the randomness in selecting unlabeled data and the noise in the pseudo-labels. 

Building on Lemma~\ref{lem1}, we demonstrate that the errors of the view-specific classifiers are expected to decrease under certain conditions. The first theorem focuses on the convergence of the iterative co-training process. 

\begin{theorem}[Co‐Training Convergence]\label{theo1}
Let $h^{(1,k)}$ and $h^{(2,k)}$ be the classifiers on views 1 and 2 after $k$ rounds of co‐training, with true error rates $\epsilon^{(1,k)}$ and $\epsilon^{(2,k)}$. Under Assumptions~\ref{ass:1} and~\ref{ass:2}, there exists
\begin{align}
    \alpha\in(0,1),\;\lambda =1-\alpha\in(0,1),\;c_{\min}\ge0
\end{align}
such that, with high probability for each $i\in\{1,2\}$ and $j\neq i$,
\begin{align}
    \mathbb{E}\bigl[\epsilon^{(i,k+1)}\bigr]\le
(1-\alpha)\,\epsilon^{(i,k)}+\alpha\,\epsilon^{(j,k)}+c_{\min},
\end{align}
and hence
\begin{align}
    \max_{i=1,2}\mathbb{E}\bigl[\epsilon^{(i,k+1)}\bigr]\le
\lambda\max_{i=1,2}\epsilon^{(i,k)}+c_{\min}<\max_{i=1,2}\epsilon^{(i,k)}.
\end{align}
Here, $c_{\min}$ accounts for irreducible noise or finite‐data effects, and in the ideal infinite‐data limit, it approaches the Bayes error of the combined views.
\end{theorem}

Theorem~\ref{theo1} extends this single-step improvement to an iterative process, showing geometric convergence of the maximum error across views towards a residual error floor ($c_{\min}$). The contraction factor ($\lambda$) dictates the speed of convergence: a smaller $\lambda$ and larger $\alpha$ lead to faster error reduction and more effective learning from the peer view, respectively. $c_{\min}$ represents the limit of what can be learned through this process, influenced by factors like inherent noise, the Bayes error of the combined views, and finite sample effects. 


\begin{figure}
    \centering
    \includegraphics[width=0.8\linewidth]{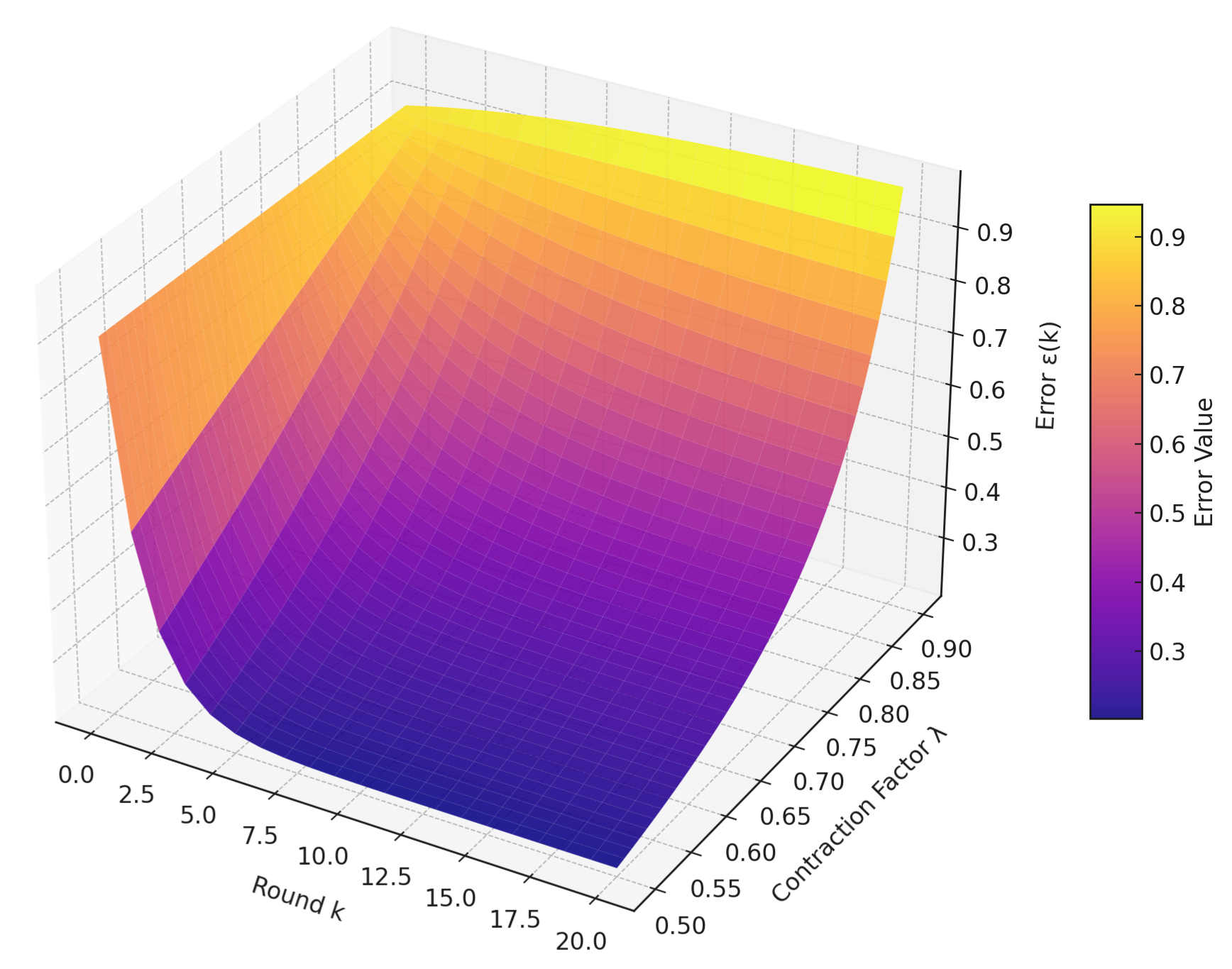}
    \caption{Error‐Contraction Surface from Simulation. It illustrates how the maximum error $\epsilon^{(k)}$ evolves over different rounds $k$ and contraction factors $\lambda$.}
    \label{fig:error-contract}
\end{figure}

This theorem also shows that both view-specific classifiers improve iteratively and converge to a small residual error under ideal co-training assumptions. At each round $k$, Lemma~\ref{lem1} ensures that retraining classifier $h^{(i,k)}$ on sufficiently many pseudo-labels from the other view $h^{(j,k)}$ yields an expected error (Eq.(7)) with mixing weight $\alpha \in(0,1)$ and a nonnegative floor $c_{\min}$ accounting for irreducible noise or finite-data effects. Whenever $\epsilon^{(j, k)}<\epsilon^{(i, k)}$, this bound guarantees $\mathbb{E}\left[\epsilon^{(i, k+1)}\right]<\epsilon^{(i, k)}$, establishing an improvement in each one-step update. By applying the same bound to both views and taking the maximum error into account, Theorem~\ref{theo1} demonstrates a geometric contraction up to the residual $c_{min}$ (Eq.(8)). This type of convergence behavior is a hallmark of successful bootstrapping algorithms. Similar analyses exist for self-training~\cite{oymak2021theoretical} and other semi-supervised methods~\cite{sheng2018convergence}; however, our result is specific to the multimodal co-training dynamics and the interplay between two classifiers. 

Figure~\ref{fig:error-contract}'s surface shows a rapid decay of error, transitioning from warm colors to cool colors, and approaches an asymptote at the irreducible error floor. A higher degree of independence or a larger accuracy gap results in a smaller $\lambda$, leading to faster convergence. As more representative unlabeled data is incorporated, the value of $c_{min}$ decreases, bringing the error floor closer to the Bayes optimum. 

\subsection{Generalization Bound for Multimodal Co-Training}
Beyond convergence, it is important to understand how well the learned models will generalize to unseen data. Before the main theorem, we introduce a proposition that clarifies the nature of the benefit derived from unlabeled data in this co-training context.

\begin{proposition}[Benefit of Unlabeled Data]\label{prop:gamma}
In multimodal co-training over $N_L$ labeled and $N_U$ unlabeled examples, the "benefit" term
\begin{align}
    \Gamma =\Gamma\Bigl(\tfrac{N_U}{N_L+N_U},d(h^{(1)},h^{(2)}),\mathrm{indep}\Bigr)
\end{align}
satisfies $\Gamma\ge 0$ and
\begin{align}
    \frac{\partial\Gamma}{\partial\bigl(N_U/(N_L+N_U)\bigr)}>0,\;
    \frac{\partial\Gamma}{\partial\,d(h^{(1)},h^{(2)})}<0,\;
    \frac{\partial\Gamma}{\partial\,\mathrm{indep}}>0.
\end{align}
Equivalently:
\begin{enumerate}[(i)]
  \item $\Gamma$ grows with the fraction of unlabeled data $\tfrac{N_U}{N_L+N_U}$.
  \item $\Gamma$ increases as the classifiers’ disagreement $d(h^{(1)},h^{(2)})$ decreases.
  \item $\Gamma$ increases with stronger conditional independence of the two views.
\end{enumerate}
\end{proposition}

Proposition~\ref{prop:gamma} introduces the benefit term $\Gamma$, which is central to our novel generalization bound. It establishes that $\Gamma$ is not merely an abstract quantity but is directly and positively influenced by the proportion of unlabeled data utilized, the degree of agreement (low $d\left(h^{(1)}, h^{(2)}\right)$) between the view-specific classifiers on this unlabeled data, and the strength of conditional independence between the views. This proposition sets the stage for understanding how co-training translates unlabeled data and inter-view consistency into improved generalization.

The proposition illustrates that unlabeled data becomes valuable only when co-training leverages the agreement between two independent views. When the classifiers $h^{(1)}$ and $h^{(2)}$ seldom disagree on the majority of unlabeled examples, their matching labels serve as a high-confidence signal. This helps to regularize the hypothesis space and reduces the risk of overfitting to the small labeled dataset. The disagreement rate $d$ serves as a measure of view alignment: a low $d$ indicates that the models capture compatible structures, which enhances $\Gamma$. In this case, $\Gamma$ quantifies the value that co-training derives from unlabeled data through multiview consistency, as determined by theoretical generalization bounds that improve as the level of disagreement among the unlabeled data decreases.

\begin{theorem}[Generalization Bound]\label{theo2}
Let $h_\theta\in\mathcal{H}$ be the model returned by multimodal co-training on $N_L$ labeled and $N_U$ unlabeled samples, where $\mathcal{H}$ has effective complexity $d_{\mathrm{eff}}$. Under Assumptions~\ref{ass:1}-\ref{ass:3}, for any $\delta\in(0,1)$, with probability at least $1-\delta$:
\begin{align}
    R(h_\theta) &\le\underbrace{\hat R_{N_L}(h_\theta)}_{\text{empirical risk}}+ C_1\sqrt{\frac{d_{\mathrm{eff}}\ln(N_L/d_{\mathrm{eff}})+\ln(1/\delta)}{N_L}} \\
    &-\underbrace{\Gamma\Bigl(\tfrac{N_U}{N_L+N_U},\mathrm{agreement},\mathrm{indep}\Bigr)}_{\substack{\text{unlabeled benefit}}}+
C_2\sqrt{\frac{\ln(1/\delta)}{N_L+N_U}},
\end{align}
where $R(h_\theta)$ is the true risk, $\hat R_{N_L}(h_\theta)$ the empirical risk on labeled data, $\Gamma\ge0$ grows with the unlabeled fraction, classifier agreement, and view independence, and $C_1,C_2$ are universal constants.  
\end{theorem}

Theorem~\ref{theo2} presents our main result in generalization theory: a PAC-style bound for multimodal co-training. Its primary novelty lies in the subtractive unlabeled benefit term ($-\Gamma(\dots)$), While prior co-training bounds~\cite{blum1998combining, dasgupta2001pac} established that co-training can reduce error, this bound offers a more granular decomposition, attributing a quantifiable benefit to the interplay of unlabeled data volume, inter-view agreement, and view independence. This contrasts with bounds where such benefits might be implicitly absorbed into complexity terms or empirical risk calculations. The term $\hat{R}_{N_L}\left(h_\theta\right)$ is the standard empirical risk on $N_L$ labeled samples, while $C_1 \sqrt{\frac{d_{\text{eff}} \ln \left(N_L / d_{\text{eff}}\right)+\ln (1 / \delta)}{N_L}}$ is a standard complexity penalty for supervised learning. The core innovation is $-\Gamma(\dots)$, a negative term that reduces the overall error bound; its magnitude, detailed in Proposition~\ref{prop:gamma}, grows with more unlabeled data, better agreement, and stronger independence, directly modeling the benefit co-training derives from unlabeled data under favorable view conditions. The final term, $C_2 \sqrt{\frac{\ln (1 / \delta)}{N_L+N_U}}$, can be interpreted as a smaller penalty related to the confidence in estimating agreement from the total pool of $N_L+N_U$ data, which also diminishes as $N_U$ grows.
 
Recent advancements in SSL generalization theory have explored various avenues, including information-theoretic bounds~\cite{chen2021generalization}, bounds for SSL under distribution shifts~\cite{saberi2023out,li2021semantic}, and bounds accounting for multimodal interactions~\cite{liang2023multimodal, liang2023quantifying, lian2022smin}. More recently, \cite{rodemann2025generalization} has proposed universal generalization bounds for reciprocal learning that aim to minimize distributional assumptions by focusing on verifiable algorithmic conditions. Theorem~\ref{theo2} makes a unique contribution by providing a PAC-style analysis that disentangles the specific contributions of unlabeled data characteristics within the co-training paradigm, offering a transparent analytical tool for understanding these interactions. It builds upon previous work~\cite{blum1998combining,dasgupta2001pac} by isolating and quantifying the benefits derived from unlabeled data as a distinct, subtractive term within the generalization bound, which is significant in providing a clear, interpretable, and structured framework for understanding generalization in multimodal co-training. Table~\ref{tab:comparative} presents a comparative analysis of this generalization bound with other relevant paradigms.

\begin{table*}[ht]
\centering
\caption{Comparative Analysis of Generalization Bounds in Co-Training and Related Semi-Supervised Learning Paradigms}
\label{tab:comparative}
\setlength{\extrarowheight}{2pt}
\setlength{\tabcolsep}{6pt}
\small  

\begin{tabularx}{\textwidth}{%
  >{\columncolor[HTML]{F8FAFD}\raggedright\arraybackslash}X  
  >{\columncolor[HTML]{F8FAFD}\raggedright\arraybackslash}X
  >{\columncolor[HTML]{F8FAFD}\raggedright\arraybackslash}X
  >{\columncolor[HTML]{F8FAFD}\raggedright\arraybackslash}X
  >{\columncolor[HTML]{F8FAFD}\raggedright\arraybackslash}X
  >{\columncolor[HTML]{F8FAFD}\raggedright\arraybackslash}X
}
\toprule
\textbf{\begin{center}Bound/Framework\end{center}}
  & \textbf{\begin{center}Key Feature\end{center}}
  & \textbf{\begin{center}Accounts for Inter-View Agreement\end{center}}
  & \textbf{\begin{center}Accounts for View Conditional Independence\end{center}}
  & \textbf{\begin{center}Handles Multimodality\end{center}}
  & \textbf{\begin{center}Primary Assumptions\end{center}} \\
\midrule
This Work (Theorem~\ref{theo2})
  & Yes ($\Gamma$ term, subtractive)
  & Yes (explicit in $\Gamma$)
  & Yes (explicit in $\Gamma$, and as our assumptions)
  & Explicitly (dual views)
  & View Sufficiency; Conditional Independence; Pseudo-label Reliability \\
\addlinespace
\cite{dasgupta2001pac}
  & Implicit (error reduction shown)
  & Yes (via disagreement rate)
  & Yes (as assumption)
  & Two-view specific
  & Conditional Independence; $\epsilon$-good classifiers \\
\addlinespace
\cite{rodemann2025generalization}
  & Focus on algorithmic conditions, not an explicit benefit term
  & N/A
  & Aims to avoid such assumptions
  & General SSL, not modality-specific
  & Verifiable algorithmic properties \\
\addlinespace
Generic SSL Rademacher Bound~\cite{bartlett2002rademacher}
  & Typically via empirical risk on unlabeled data or complexity modulation
  & Typically via consistency regularization
  & Cluster, manifold, etc.
  & General SSL
  & Depends on method (cluster assumption; manifold assumption; low-density separation) \\
\bottomrule
\end{tabularx}
\end{table*}

\begin{corollary}[Effect of Unlabeled Sample Size]\label{cor:NU}
Under Assumptions~\ref{ass:1}-\ref{ass:3}, as $N_U$ grows with $N_L$ fixed:
\begin{align}
    \Gamma\bigl(\tfrac{N_U}{N_L+N_U},\text{agreement},\text{indep}\bigr)\uparrow
\;\text{and}\;
C_2\sqrt{\frac{\ln(1/\delta)}{N_L+N_U}}\downarrow.
\end{align}
Consequently, the overall bound in Theorem~\ref{theo2} tightens monotonically in $N_U$.
\end{corollary}

This corollary highlights the data-efficiency benefits of co-training. As the pool of unlabeled examples increases, the benefit term $\Gamma$ grows because there are more opportunities for the two views to reach a high-confidence agreement. At the same time, the confidence penalty $C_2 \sqrt{\ln(1/\delta) / (N_L + N_U)}$ decreases. This dual effect demonstrates a powerful leveraging of easily accessible unlabeled resources to tighten the generalization bound, a key desideratum in low-resource settings. 

\begin{corollary}[Effect of View Independence and Agreement]\label{cor:views}
Under Assumptions~\ref{ass:1}-\ref{ass:3}, the benefit $\Gamma$ increases whenever:
\begin{enumerate}[(i)]
    \item The views show stronger conditional independence.
    \item The disagreement rate $d(h^{(1)},h^{(2)})$ on $\mathcal{U}$ decreases.
\end{enumerate}
Equivalently, greater view independence and higher empirical agreement both amplify $\Gamma$.
\end{corollary}

\begin{figure}[!ht]
    \centering
    \includegraphics[width=0.8\linewidth]{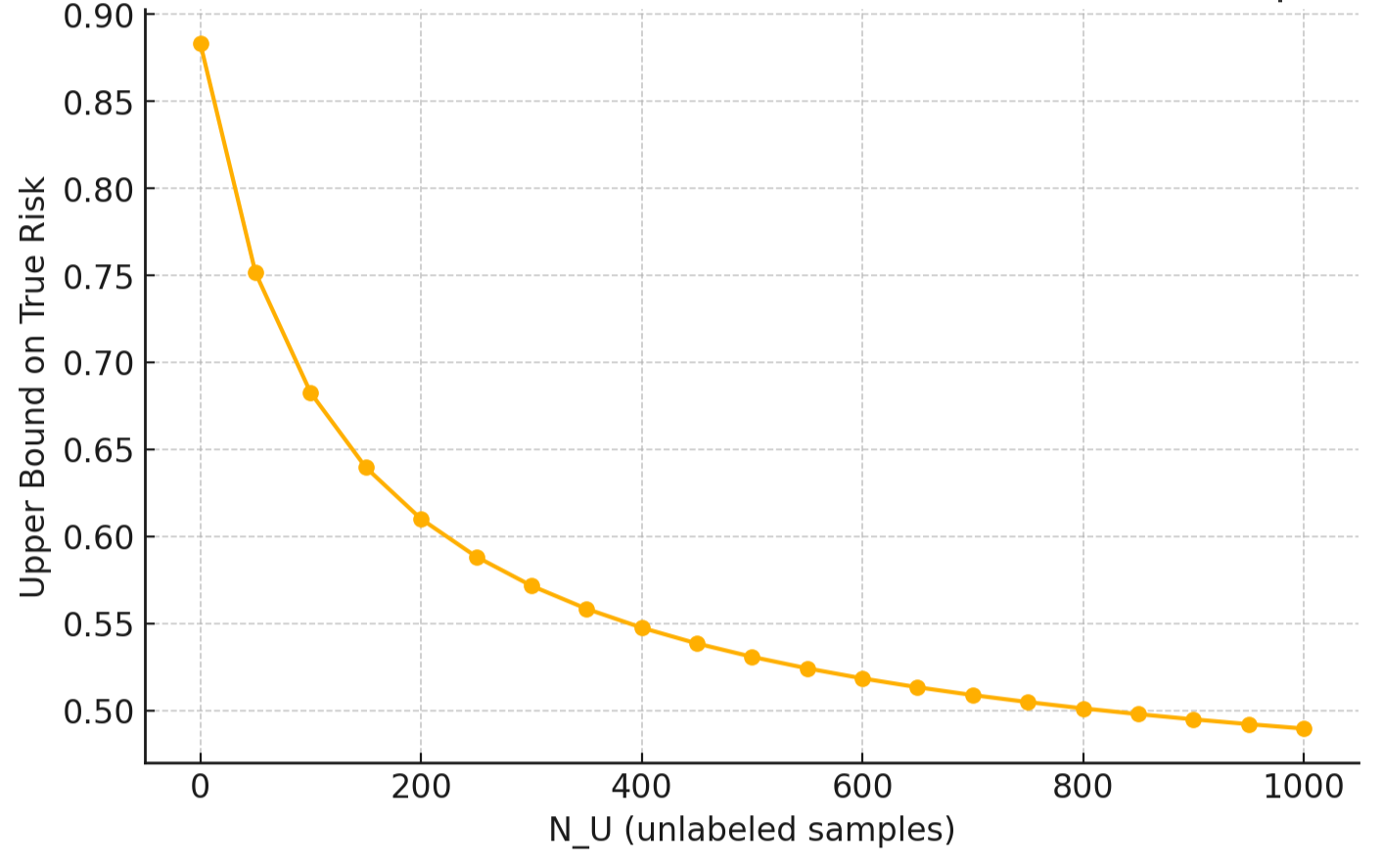}
    \caption{Generalization Bound vs. Number of Unlabeled Samples from Simulation. We fix a small empirical risk and hypothetical constants, then let $\Gamma$ grow proportionally to the unlabeled data fraction $\frac{N_U}{N_L+N_U}$.}
    \label{fig:bound-sim}
\end{figure}

\begin{figure}[!ht]
    \centering
    \includegraphics[width=0.8\linewidth]{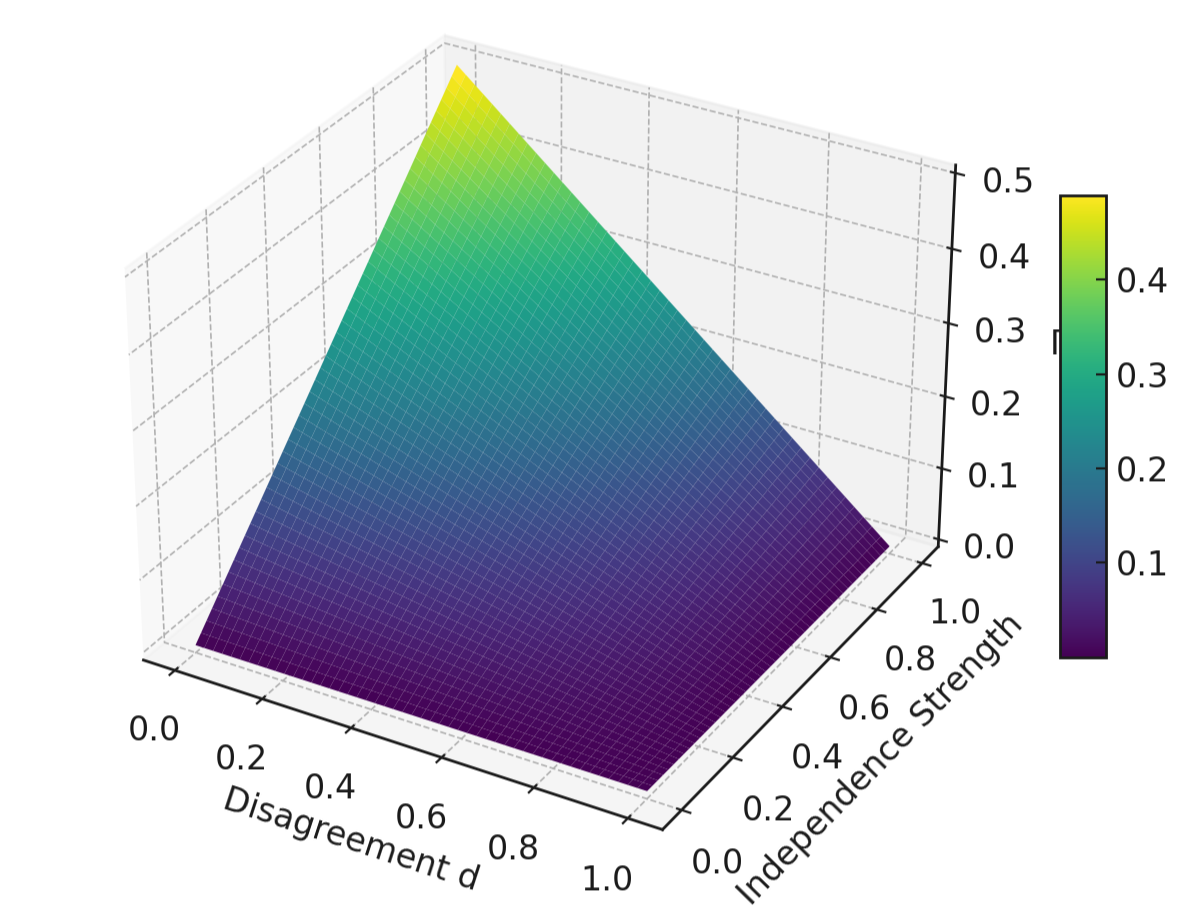}
    \caption{Benefit $\Gamma$ vs. Disagreement and Independence from Simulation. With a fixed unlabeled data fraction, we visualize $\Gamma = \text{frac}\cdot (1-d) \cdot \text{indep}$. The surface peaks when views are maximally independent and fully agree.}
    \label{fig:gamma-sim}
\end{figure}

This corollary highlights the qualitative aspects crucial for co-training success. Stronger conditional independence ensures that agreement is not spurious (i.e., due to highly correlated views merely echoing each other's biases) but genuinely reflects shared underlying truth about the data. Lower disagreement (higher empirical agreement) on unlabeled data, when coupled with this independence, provides robust corroborating evidence for pseudo-labels, directly amplifying the benefit term. This emphasizes the importance of careful view construction or feature engineering to foster both competence (sufficiency) and diversity (leading to conditional independence) in the views. Figures~\ref{fig:bound-sim} and~\ref{fig:gamma-sim} visually illustrate these dependencies, showing how the generalization bound tightens with more unlabeled samples and how the benefit term increases with lower disagreement and stronger independence, respectively, based on simulated data reflecting the relationships in Theorem~\ref{theo2} and its corollaries. 

\section{Conclusion}
This paper establishes a rigorous theoretical basis for our proposed multimodal co-training framework. We have demonstrated its capacity for reliable convergence and derived a novel generalization bound that specifically quantifies how unlabeled data, inter-view agreement, and view independence contribute to enhanced generalization performance under data scarcity and potential distribution shifts. Through thorough theoretical analysis, we demonstrate that co-training can reliably converge, with classification errors decreasing iteratively as long as the modality-specific views remain sufficiently independent and consistently agree on the unlabeled data. The generalization bound we derived quantifies the benefits of using unlabeled multimodal data, offering concrete guidance for practitioners looking to enhance performance through data-efficient training strategies. 

Our findings offer more than theoretical validation, providing concrete and actionable insights for practitioners seeking to develop data-efficient and robust AI systems. The clear dependence of our generalization bound on measurable quantities (unlabeled data fraction, agreement rates, proxies for independence) offers guidance for optimizing co-training strategies. 

\subsection{Limitation and Future Work}
Our work is not without limitations. The relaxation of the conditional independence assumption within our framework, potentially incorporating concepts such as ``view expansion"~\cite{balcan2004co} to develop bounds applicable to a wider range of real-world scenarios where this strong assumption may not fully hold. Future work could investigate adaptive mechanisms for the agreement loss $\mathcal{L}_{\text{agree}}$ or the label-expansion budget in Algorithm~\ref{alg:multimodal_cotraining_concise}, informed by the evolving disagreement rates or confidence scores during training, to optimize the learning process further. Also, future research could extend the analysis to handle more than two modalities and explore the interplay with modern large-scale pre-trained models when used as view-specific feature extractors, thereby leveraging their powerful representations within the co-training paradigm. Moreover, empirically validating the tightness of the derived generalization bound and its components on diverse multimodal datasets that experience various forms of distribution shifts can further bridge the gap between theory and practice. Additionally, developing stopping rules for co-training based on anytime valid generalization bounds would provide practical criteria for terminating training while ensuring performance guarantees, making this approach valuable.

\bibliographystyle{unsrt}  
\bibliography{references}  

\newpage
\appendix
\onecolumn
\section{More Related Work}
\subsection{Few-Shot and Zero-Shot Learning (FSL/ZSL)}
FSL and ZSL aim to facilitate learning with very few or no labeled examples per class. While these methods are promising in limited-data scenarios, they encounter significant challenges in high-dimensional or multimodal settings. For instance, ZSL often struggles to generalize with domain-specific data, leading to lower accuracy and potential bias. FSL, on the other hand, can be sensitive to the selection of support examples or the design of prompts, impacting its reliability~\cite{chamieh2024llms}. Challenges of multimodal scenarios, such as few-shot action recognition using video data, include effectively extracting spatiotemporal features and learning generalizable representations from limited sample sizes~\cite {liu2025boosting,tang2024semantic}. High intra-class variance, common in real-world data, poses particular challenges and often results in overfitting when sample sizes are scarce. Additionally, traditional ZSL typically assumes a single semantic attribute vector per class, which is inadequate for categories with diverse appearances or multiple distinct modes. For example, different plumages of a bird species may not be well represented by a ``one-size-fits-all" attribute vector, leading to too general and non-discriminative attributes~\cite{wanyan2024comprehensive}.

The proposed multimodal co-training framework presents a novel approach that utilizes at least two modalities and leverages unlabeled data. This design accommodates various aspects of the data. For example, if one view captures textual descriptions and another captures visual appearances, co-training can address the limitations of relying on a ``single attribute vector." This method provides richer representations, enabling more effective handling of intra-class variance than many FSL or ZSL methods, especially when a wealth of unlabeled multimodal data is available.

\subsection{Multimodal Learning}
Multimodal learning integrates information from various data types, such as text, images, and audio~\cite{li2024multimodal}. Many supervised approaches are ``fusion-heavy," meaning they combine features or decisions from different modalities at different stages, including early, intermediate, or late fusion~\cite{xu2013survey}. Although these methods are effective with plenty of labeled data, they usually require large amounts of labeled multimodal examples to learn valuable cross-modal interactions and create robust fused representations~\cite{yu2025review}. Several challenges arise in this process, such as ensuring that the features are semantically aligned before fusion, managing the variability and reliability differences between modalities, handling high-dimensional data, and achieving optimal fusion of views. When labeled data is limited, the features learned for each modality can be noisy or unrepresentative, resulting in suboptimal fusion outcomes.

The proposed multimodal co-training framework introduces a unique approach to data processing. Instead of solely relying on the direct integration of representations obtained from limited labeled data, it utilizes unlabeled data and the principle of cross-view consistency~\cite{li2024multimodal}. This method regularizes and enhances individual learners' specific views before any final integration takes place, resulting in more refined features. The agreement mechanism is an implicit form of learned integration driven by unlabeled data. This positions co-training as a more efficient alternative or an essential complement to supervised integration.

\subsection{Semi-supervised learning (SSL) and Co-Training}
SSL methods utilize both labeled and unlabeled data for training. The co-training paradigm, introduced by~\cite{blum1998combining}, is especially effective in multimodal settings. It is based on two main assumptions: (1) multiple "views" or modalities are independently sufficient for classification, and (2) these views are conditionally independent given the class label. Classifiers trained on their respective views iteratively assign pseudo-labels to confident examples from the unlabeled data, allowing them to train each other. This approach enables the models to effectively "teach" one another, thereby expanding the training set by leveraging information from unlabeled data.

While the foundational principles of co-training are well-known, there remains a lack of rigorous analyses, particularly concerning convergence and generalization bounds in multimodal semi-supervised learning. This gap is especially evident when examining aspects such as the adequacy of different views, conditional independence, and the role of unlabeled data in refining these bounds. This paper aims to deepen the theoretical understanding of these concepts by providing a more robust mathematical framework for applying co-training in complex, multimodal, and data-scarce settings.

\subsection{Open-World Generalization and Distribution Shift}
Open-world deployment requires models to effectively generalize under non-stationary data distributions, where concept drift or distribution shift can significantly degrade performance~\cite{lu2018learning}. For instance, misinformation campaigns continuously evolve in narratives and styles, which can cause static models to struggle. While some approaches utilize drift detection mechanisms to initiate model updates, it is important to understand the robustness of different learning paradigms in response to such shifts~\cite{bayram2022concept}. Recent Test-Time Adaptation (TTA) techniques aim to adjust models to ongoing shifts~\cite{wu2025multi,dong2025advances}, but the challenge remains significant, with even large foundation models sometimes faltering under such conditions. 

The co-training mechanism in the proposed framework allows for a degree of passive adaptation to evolving data streams. This theoretical framework examines the generalization bounds that incorporate agreement on unlabeled data, providing insights into how co-training can maintain performance. If unlabeled data is drawn from the current, potentially shifted distribution, enforcing consistency on this data can help the model stay aligned with changes. This paper offers a foundational understanding of co-training behavior in non-stationary environments, which is important for analyzing more complex adaptive systems. A key consideration is the robustness of the derived bounds in the face of variations introduced by non-stationarity.

\section{Proof}
\subsection{Proof of Lemma~\ref{lem1}}
\begin{proof}\label{proof:lemma}
    Let $D$ denote the true data distribution over $(x, y)$, and let $D_{\mathrm{pl}}$ be the distribution over ( $x, y$ ) obtained by drawing $x \sim D$ and assigning it the pseudo-label $\bar{y}=$ $h^{(2)}(x)$. For any mixing weight $\alpha \in(0,1]$, define the mixture
    $$
    D_{\text {mix }}=(1-\alpha) D+\alpha D_{\mathrm{pl}}
    $$
    Under Assumptions~\ref{ass:1} and \ref{ass:2}(ii), the Bayes error on $D_{\mathrm{pl}}$ equals $\backslash \mathrm{eps}^{(2)}$. Therefore, the Bayes risk on the mixture satisfies
    $$
    R_{D_{\text {mix }}}^* \leq(1-\alpha) \epsilon^{(1)}+\alpha \epsilon^{(2)}.
    $$
    Retraining $h^{(1)}$ on $N$ i.i.d. samples from $D_{\text{mix}}$ produces a classifier whose expected error satisfies the standard excess-risk decomposition
    $$
    \mathbb{E}\left[\epsilon^{\prime(1)}\right] \leq R_{D_{\operatorname{mix}}}^*+\xi_N,
    $$
    where $\xi_N \rightarrow 0$ as $N \rightarrow \infty$ by uniform convergence (e.g., VC or Rademacher complexity bounds). Combining the above,
    $$
    \mathbb{E}\left[\epsilon^{\prime(1)}\right] \leq(1-\alpha) \epsilon^{(1)}+\alpha \epsilon^{(2)}+\xi_N
    $$

    Since $\epsilon^{(2)}<\epsilon^{(1)}$, we can choose $N$ large enough that
    $\xi_N \leq \frac{\alpha}{2}\left(\epsilon^{(1)}-\epsilon^{(2)}\right)$. Then, 
    $$
    (1-\alpha) \epsilon^{(1)}+\alpha \epsilon^{(2)}+\xi_N<(1-\alpha) \epsilon^{(1)}+\alpha \epsilon^{(2)}+\frac{\alpha}{2}\left(\epsilon^{(1)}-\epsilon^{(2)}\right)=\epsilon^{(1)},
    $$
    establishing
    $\mathbb{E}\left[\epsilon^{\prime(1)}\right]<\epsilon^{(1)}$. This completes the proof of Lemma~\ref{lem1}.
\end{proof}

\subsection{Proof of Theorem~\ref{theo1}}
\begin{proof}\label{proof:theo1}
    By Lemma~\ref{lem1}, when classifier $h^{(i, k)}$ is retrained on a sufficiently large, high-confidence pseudo-labeled set provided by $h^{(j, k)}$, its expected error satisfies
    $$
    \mathbb{E}\left[\epsilon^{(i, k+1)}\right] \leq(1-\alpha) \epsilon^{(i, k)}+\alpha \epsilon^{(j, k)},
    $$
    for some mixing weight $\alpha>0$ determined by the fraction and reliability of pseudo-labels.
    In practice, an irreducible floor $c_{\min} \geq 0$ accounts for Bayes-error and finite-sample effects, yielding exactly inequality (7).
    Define the worst-case error at round $k$ as
    $$
    \epsilon_{\max }^{(k)}=\max \left\{\epsilon^{(1, k)}, \epsilon^{(2, k)}\right\}
    $$
    Applying (7) for both $i=1$ and $i=2$ and then taking the maximum gives
    $$
    \max _i \mathbb{E}\left[\epsilon^{(i, k+1)}\right] \leq \max _i\left[(1-\alpha) \epsilon^{(i, k)}+\alpha \epsilon^{(j, k)}+c_{\min}\right] \leq(1-\alpha) \epsilon_{\max }^{(k)}+\alpha \epsilon_{\max}^{(k)}+c_{\min}=\epsilon_{\max}^{(k)}+c_{\min}
    $$
    Setting $\lambda=1-\alpha$ yields the geometric-contraction form
    $$
    \max _i \mathbb{E}\left[\epsilon^{(i, k+1)}\right] \leq \lambda \epsilon_{\max}^{(k)}+c_{\min}
    $$
    as claimed. Whenever $\epsilon^{(j,k)}<\epsilon^{(i,k)}$ (ensured by Assumption~\ref{ass:2}(i)), each convex combination $(1-\alpha)\epsilon^{(i,k)} + \alpha\epsilon^{(j,k)}$ is strictly less than $\epsilon^{(i,k)}$. Thus, if 
    $$
    c_{\min}<(1-\lambda) \epsilon_{\max}^{(k)}=\alpha \epsilon_{\max}^{(k)},
    $$
    then, 
    $$
    \lambda\max_{i=1,2}\epsilon^{(i,k)}+c_{\min}<\max_{i=1,2}\epsilon^{(i,k)},
    $$
    which establishes a strict geometric decrease in the worst-case error each round. Together, these show that under the standard co-training assumptions, the maximum of the two view-errors shrinks geometrically toward the irreducible floor $c_{\min}$. This completes the proof of Theorem~\ref{theo1}
\end{proof}

\subsection{Proof of Proposition~\ref{prop:gamma}}
\begin{proof}
    By construction (see the mixture-error argument in Proof~\ref{proof:lemma}), enforcing agreement on pseudo-labels can only decrease the Bayes risk on the retraining mixture compared to training purely on labeled data. Hence, the subtractive term $\Gamma$ in the generalization bound is always nonnegative.
    Let
    $$
    \eta=\frac{N_U}{N_L+N_U}
    $$
    be the weight on the pseudo-labeled component in the retraining mixture $D_{\text {mix }}=(1-\eta) D+\eta D_{\text {pl }}$. A larger $\eta$ places more emphasis on the pseudo-labels, which yields strictly greater error reduction under Assumptions~\ref{ass:1} and \ref{ass:2}. Thus,
    $$
    \frac{\partial \Gamma}{\partial \eta} \;>\;0 \; \text{(Monotonicity in the unlabeled fraction)}. 
    $$
    Define
    $$
    d\left(h^{(1)}, h^{(2)}\right)=\operatorname{Pr}_{x \sim P}\left[h^{(1)}(x) \neq h^{(2)}(x)\right]
    $$
    on the unlabeled pool. A higher disagreement indicates that the two views disagree more frequently on pseudolabels, thereby reducing their joint reliability. Consequently, the co-training benefit $\Gamma$ decreases as $d$ increases. Thus, $$\frac{\partial \Gamma}{\partial\,d\bigl(h^{(1)},h^{(2)}\bigr)} \;<\;0 \; \text{(Monotonicity in view disagreement)}.$$
    Finally, stronger conditional-independence, i.e. a tighter factorization $P\left(x^{(1)}, x^{(2)} \mid y\right)=P\left(x^{(1)} \mid\right.$ y) $P\left(x^{(2)} \mid y\right)$ guarantees that agreement between the two views is more likely to reflect the true label rather than correlated errors. Hence, as the independence parameter increases, so does the quality of pseudo-labels, and thus
    $$
    \frac{\partial \Gamma}{\partial\,\mathrm{indep}} \;>\;0 \; \text{(Monotonicity in conditional independence)}.
    $$
    Since each partial derivative has the stated sign, the three equivalent monotonicity claims follow immediately. This completes the proof of Proposition~\ref{prop:gamma}.
\end{proof}

\subsection{Proof of Theorem~\ref{theo2}}
\begin{proof}
    By standard VC/Rademacher arguments, with probability at least $1-\frac{\delta}{2}$ over the draw of $\mathcal{L}$, every $h \in \mathcal{H}$ satisfies 
    \begin{align*}
        R(h) \leq \widehat{R}_{N_L}(h)+C_1 \sqrt{\frac{d_{\mathrm{eff}} \ln \left(N_L / d_{\mathrm{eff}}\right)+\ln (2 / \delta)}{N_L}}.
    \end{align*}
    Here, $d_{\text{eff}}$ is either the VC dimension or a Rademacher-based measure of $\mathcal{H}$, and $C_1$ is an absolute constant. By design, our co-training algorithm enforces that $h$ achieves empirical agreement on the unlabeled pool: it either pseudo-labels or directly penalizes discrepancies via $L_{\text{agree}}$. Let $A(h)=\frac{1}{N_U} \sum_{j=1}^{N_U} 1\left\{h^{(1)}\left(u_j\right) \neq h^{(2)}\left(u_j\right)\right\}$ be the empirical disagreement rate, and write $d\left(h^{(1)}, h^{(2)}\right)$ for its true counterpart. Proposition~\ref{prop:gamma} shows that enforcing low $A(h)$ under conditional independence yields a nonnegative benefit $\Gamma\left(\frac{N_l}{N_l+N_U}, A(h), \text {indep}\right)$ such that, on the event $\left\{A(h) \approx d\left(h^{(1)}, h^{(2)}\right)\right\}$, one can subtract $\Gamma$ from the usual risk. Formally, 
    \begin{align*}
        R(h) \leq \widehat{R}_{N_L}(h)-\Gamma\left(\frac{N_U}{N_L+N_U}, A(h), \text {indep}\right)+\Delta(h),
    \end{align*}
    where $\Delta(h)$ captures any slack from finite unlabeled-sample estimation of $d$. By a Hoeffding bound on the $N_U$ independent indicator variables in $A(h)$, with probability at least $1-\frac{\delta}{2}$, uniformly over $\mathcal{H}$: $\left|A(h)-d\left(h^{(1)}, h^{(2)}\right)\right| \leq C_2 \sqrt{\frac{\ln (2 / \delta)}{N_U}}$ for some absolute constant $C_2$. Hence, we can absorb $\Delta(h)$ into a single term $C_2 \sqrt{\frac{\ln (2 / \delta)}{N_L+N_U}}$ since $N_L$ and $N_U$ both contribute to our overall confidence budget. Taking a union bound over the two $1-\frac{\delta}{2}$ events, we get with probability $\geq 1-\delta$ that both Eq.(11) and (12) hold and that $A(h)$ is close to $d\left(h^{(1)}, h^{(2)}\right)$. Substituting the concentration of $A(h)$ into Eq.(12) and combining with $(A)$ yields
    \begin{align*}
        R(h) &\leq \widehat{R}_{N_L}(h)+C_1 \sqrt{\frac{d_{\text{eff}} \ln \left(N_L / d_{\text{eff}}\right)+\ln (2 / \delta)}{N_L}} \\&-\Gamma\left(\frac{N_U}{N_L+N_U}, \text{agreement, indep}\right)+C_2 \sqrt{\frac{\ln (2 / \delta)}{N_L+N_U}}.
    \end{align*}
    Renaming constants and replacing $\ln (2 / \delta)$ by $\ln (1 / \delta)$ up to constant factors completes the stated bound. This completes the proof of Theorem~\ref{theo2}.
\end{proof}

\subsection{Proof of Corollary~\ref{cor:NU}}
\begin{proof}
    Recall the bound from Theorem~\ref{theo2} (Eq.(11) and (12)), and from Proposition~\ref{prop:gamma}, $\Gamma$ is increasing in the fraction $\frac{N_U}{N_L+N_U}$. As $N_U$ grows with $N_L$ fixed, this fraction increases monotonically, so $\Gamma$ strictly increases. Further, the term $C_2\sqrt{\frac{\ln(1/\delta)}{N_L+N_U}}$ is a strictly decreasing function of the total sample size $N_L+N_U$. Hence, as $N_U$ increases, this penalty shrinks. Combining these two monotonicities shows that the RHS of the generalization bound tightens as $N_U$ grows. This completes the proof of Corollary~\ref{cor:NU}.
\end{proof}

\subsection{Proof of Corollary~\ref{cor:views}}
\begin{proof}
    From Theorem~\ref{theo2}, we have
    $$
    R\left(h_\theta\right) \leq \widehat{R}_{N_L}\left(h_\theta\right)+C\left(N_L, d_{\mathrm{eff}}, \delta\right)-\Gamma\left(\eta, d\left(h^{(1)}, h^{(2)}\right), \text{indep}\right)+C^{\prime}\left(N_L+N_U, \delta\right)
    $$
    Here $\Gamma\geq 0$ by Proposition~\ref{prop:gamma}, and the two residual terms $C$ and $C^{\prime}$ do not depend on view-agreement or independence.
    By Proposition~\ref{prop:gamma},
    $$
    \frac{\partial \Gamma}{\partial d\left(h^{(1)}, h^{(2)}\right)}<0 \quad \Longrightarrow \quad \frac{\partial[-\Gamma]}{\partial d\left(h^{(1)}, h^{(2)}\right)}>0
    $$
    Thus, as the disagreement $d\left(h^{(1)}, h^{(2)}\right)$ decreases (i.e., agreement increases), $-\Gamma$ strictly decreases, making the RHS of the bound strictly smaller.
    Similarly, Proposition~\ref{prop:gamma} gives
    $$
    \frac{\partial \Gamma}{\partial \text{indep}}>0 \Longrightarrow \frac{\partial[-\Gamma]}{\partial \text{indep}}<0.
    $$
    Hence, as the conditional-independence assumption strengthens (i.e., the factorization $P\left(x^{(1)}, x^{(2)} \mid\right.$ $y)=P\left(x^{(1)} \mid y\right) P\left(x^{(2)} \mid y\right)$ becomes tighter), $-\Gamma$ again decreases, tightening the bound.
    Neither $C$ nor $C^{\prime}$ depends on view-agreement or independence, so the only terms affected by these factors are $\pm \Gamma$. Since increased agreement and stronger independence both strictly increase $\Gamma$, they strictly decrease the RHS of the generalization bound. This completes the proof of Corollary~\ref{cor:views}.
\end{proof}


\end{document}